\documentclass[10pt, conference, compsocconf]{IEEEtran}
\usepackage{hyperref}
\usepackage{url}
\usepackage{graphicx}
\usepackage{amsmath}
\usepackage{multirow}
\usepackage{caption}
\usepackage{subcaption}
\begin{document}

\title{Revealing Fundamental Physics from the Daya Bay Neutrino Experiment using Deep Neural Networks}
\author{\IEEEauthorblockN{Evan Racah\IEEEauthorrefmark{1},
Seyoon Ko\IEEEauthorrefmark{2},
Peter Sadowski\IEEEauthorrefmark{3}, 
Wahid Bhimji\IEEEauthorrefmark{1},
Craig Tull\IEEEauthorrefmark{1},
Sang-Yun Oh \IEEEauthorrefmark{1}\IEEEauthorrefmark{4},
Pierre Baldi\IEEEauthorrefmark{3},
Prabhat\IEEEauthorrefmark{1}
}
\IEEEauthorblockA{\IEEEauthorrefmark{1}
Lawrence Berkeley Lab,
Berkeley, CA\\ 
Email: eracah@lbl.gov
}

\IEEEauthorblockA{\IEEEauthorrefmark{2} 
Seoul National University, Seoul 151-747, Republic of Korea 
}
\IEEEauthorblockA{\IEEEauthorrefmark{3}
University of California, Irvine, CA, USA
}
\IEEEauthorblockA{\IEEEauthorrefmark{4}
University of California, Santa Barbara, CA, USA 
}}


\maketitle
\begin{abstract}
Experiments in particle physics produce enormous quantities of data that must be analyzed and interpreted by teams of physicists. This analysis is often exploratory, where scientists are unable to enumerate the possible types of signal prior to performing the experiment. Thus, tools for summarizing, clustering, visualizing and classifying high-dimensional data are essential. In this work, we show that meaningful physical content can be revealed by transforming the raw data into a \emph{learned} high-level representation using deep neural networks, with measurements taken at the Daya Bay Neutrino Experiment as a case study. We further show how convolutional deep neural networks can provide an effective classification filter with greater than 97\% accuracy across different classes of physics events, significantly better than other machine learning approaches. 
\end{abstract}

\section{Introduction} 
The analysis of experimental data in science can be an exploratory process, where researchers do not know beforehand what they expect to observe. This is particularly true in particle physics, where extraordinarily complex detectors are used to probe the fundamental nature of the universe. These detectors collect petabytes or even exabytes of data in order to observe relatively rare events. Analyzing this data can be a laborious process that requires researchers to carefully separate and interpret different sources of signal and noise.

The Daya Bay Reactor Neutrino Experiment is designed to study anti-neutrinos produced by the Daya Bay and Ling Ao nuclear power plants. The experiment has successfully produced many important physics results 
\cite{dayabay1,dayabay2,dayabay4,dayabay5,dayabay6}  but these required significant effort to identify and explain the multiple sources of noise, not all of which were expected. For example, it was found after initial data collection that a small number of the photomultiplier tubes used in the detectors spontaneously emitted light due to discharge within their base, causing so-called ``flasher'' events. Identifying and accounting for these flashers and other unexpected factors was critical for isolating the rare antineutrino decay events. To speed up scientific research, physicists would greatly benefit from automated analyses to summarize, cluster, and visualize their data, in order to build an intuitive grasp of its structure and quickly identify flasher-like problems.

Visualization and clustering are two of the primary ways that researchers use to explore their data. This requires transforming high-dimensional data (such as an image) into a 2-D or 3-D space. One common method for doing this is principle component analysis (PCA), but PCA is linear and unable to effectively compress data that lives on a complex manifold, such as natural images. Neural networks, on the other hand, have the capacity to represent very complex transformations~\cite{hornik_multilayer_1989}. Moreover, these transformations can be learned given a sufficient amount of data. In particular, deep learning with many-layered neural networks has proven to be an effective approach to learning useful representations for a variety of application domains, such as computer vision and speech recognition \cite{alexnet}, \cite{deep_speech}. Thus, it may provide new ways for physicists to explore their high-dimensional data. Each layer of a deep feed-forward neural network computes a different non-linear representation of the input; performing exploratory data analysis on these high-level representations may be more fruitful than performing the same analysis on the raw data. Furthermore, learned representations can easily be combined with existing tools for summarizing, clustering, visualizing and classifying data.

In this work, we learn and visualize high-level representations of the particle-detector data acquired by the Daya Bay Experiment. These representations are learned using both unsupervised and supervised neural network architectures.

\section{Related Work}

Finding high-level representations of raw data is a common problem in many fields. For example, embeddings in natural language processing attempt to find a compressed vector representation for words, sentences, or paragraphs where each dimension roughly corresponds to some latent feature and distance in the embedding corresponds to semantic distance (e.g. \cite{DBLP:journals/corr/LeM14,DBLP:journals/corr/KirosZSZTUF15}). 

In addition, for natural images, extracting features and visualizing a low-dimensional manifold using autoencoders is another common application \cite{kingma2013auto}. These efforts usually are applied to well-defined datasets, such as MNIST, face image datasets and SVHN. 

In the realm of scientific data, chemical fingerprinting is a method for representing small-molecule structures as vectors~\cite{doi:10.1021/ci400187y,DBLP:journals/corr/DuvenaudMAGHAA15}. These representations are usually engineered to capture relevant features in the data, but an increasingly-common approach is to \emph{learn} new representations from the data itself, in either a supervised or unsupervised manner. Deep neural network architectures provide a flexible framework for learning these representations. 


Furthermore, deep learning has already been successfully applied to problems in particle physics. For example, Baldi et. al. showed that deep neural networks could improve exotic particle searches and showed that learned high-level features outperform those engineered by physicists~\cite{baldi_searching_2014}. Others have applied deep neural networks to the problem of classifying particle jets from low-level detector data, using convolutional architectures and treating the detector data as images~\cite{deOliveira}. 

However, these efforts have focused on supervised learning with simulated data; to the best of our knowledge, deep learning has not been used to perform unsupervised exploratory data analysis, directly on the raw detector measurements, with the goal of uncovering unexpected sources of signal and background noise.

\section{Data}
A Daya Bay Antineutrino Detector (AD) consists of 192 photomultiplier tubes (PMTs) arranged in a cylinder 8 PMTs high and with a 24 PMT circumference \cite{dayabay2}. The data we use for our study is the the value of the charge deposit of each of the PMTs in the cylinder unwrapped into a 2D (8 ``ring'' x 24 ``column'') 
array of floats. Each example is the 8x24 array for a particular event that set off a trigger to be captured. 

For the supervised part of this analysis, and for visualizing the unsupervised results, we employ labels determined by the physicists from their features and threshold criteria. For full details on these selections see \cite{dayabay2}. They label five types of events: ``muon'', ``flasher'', ``IBD prompt'', ``IBD delay'' and a default label of ``other'' is applied to all other events. 
For ``muon'' and ``flasher'' events we apply the physics selection on derived quantities held in the original data before producing our reduced data samples. 
\emph{Inverse Beta Decay} (``IBD'') labels correspond to antineutrino events that are the desired physics of interest and occur substantially less frequently than other event types. Many stages of fairly complex analysis are used by the physicists to select these~\cite{dayabay2}. Therefore, we do not reapply that selection but instead use an index output from their analyses to tag events.

Muon labelled events are relatively straightforward to cluster or learn, while flasher and IBD events involve non-linear functions and complex transformations. Furthermore, the physicists' selections for these events make use of some information that is not available to our analysis, such as times between events and with respect to external muon detectors. 
\section{Methods} 
Given a set of detector images, we aim to find a vector representation, $V \in R^n$ of each image, where n corresponds to the number of features to be learned. The features are task-specific --- they are optimized either for class-prediction or reconstruction --- but in both cases we expect the learned representation to capture high-level information about the data. By transforming the raw data into these high-level representations, we aim to provide physicists with more interpretable clusterings and visualizations, so that they may uncover unexpected sources of signal and background.

To learn new representations, we use both supervised and unsupervised convolutional neural networks. These methods are described in more detail below. As a qualitative assessment of the learned representations, we use t-Distributed Stochastic Neighbor Embedding (t-SNE) \cite{tsne}, which maps n-dimensional data to 2 or 3 dimensions and makes sure points close together in the high n-dimensional space are also close together in the lower dimensional embedding. 
\subsection{Supervised Learning with Convolutional Neural Networks}
A convolutional neural network (CNN) is a particular neural network architecture that captures our intuition about local structure and translational invariance in images \cite{alexnet}. We employ CNNs in this work because the data captured by the antineutrino detectors are essentially 2-D images. Most CNNs have several convolutional and pooling layers followed by one or more fully connected layers that use the features learned from those layers to perform typical classification or regression tasks.
\subsection{Unsupervised Learning with Convolutional Autoencoders}
An autoencoder \cite{greedy, learningdeep} is a neural network where the target output is exactly the input. It usually consists of an encoder, which consists of one or more layers that transform the input into a feature vector at the output of the middle layer (often called bottleneck layer or hidden layer), and a decoder, which usually contains several layers that attempt to reconstruct the hidden layer output back to the input. When the autoencoder architecture includes a hidden layer output with dimensionality smaller than that of the input (undercomplete), it must learn how to compress and reconstruct examples from the training data. It has been shown that undercomplete autoencoders are equivalent to nonlinear PCA \cite{kramer1991nonlinear}. In addition, there exist autoencoders that have hidden layer ouputs of higher dimension than the those of the inputs (overcomplete) that use other constraints to prevent the network from learning an identity function \cite{ kingma2013auto, vincent2008extracting, rifai2011contractive}. We use undercomplete autoencoders due to their simplicity and as an exploratory first step to see if we can indeed extract low dimensional features from this sensor data, while still taking nonlinearity into account.


A convolutional autoencoder is an autoencoding architecture that includes convolutional layers. The encoding portion typically consists of convolutional and max-pooling layers followed by fully-connected hidden layers (including a ``bottleneck'' layer) and then deconvolutional (and unpooling) layers, usually one for each convolutional and pooling layer~\cite{dumoulin2016guide}. 

While some authors~\cite{zhao2015stacked} have shown success with using deconvolutional and unpooling layers in reconstruction, we solely use transposed convolutional layers due to software constraints. Moreover, there has been work with convolutional generative models that shows success in using just fractionally strided convolutional layers and no unpooling layers~\cite{radford2015unsupervised}.

\section{Implementation} 

We performed our analysis on Edison and Cori, two Cray XC computing systems at the National Energy Research Scientific Computing Center (NERSC).


\subsection{Preprocessing}
For training and testing data, we used an equal number of examples from each of the five physics classes.
Because the muon charge deposit values are much higher than some of the other events' charge deposits, we apply a natural log transform to each value in the 8x24 image. 
For the supervised CNN, we also cyclically permute the columns, so the column containing the largest valued element in the entire array is in the center (12th column). This is done to prevent areas of interest from being located on the edges of the array (given the data is an array from an unwrapped cylinder).

\subsection{Supervised Learning with CNN}
    To help examine if there were learnable patterns in the data, we implemented a supervised convolutional neural net. The architecture of the CNN is specified in Table \ref{table:convnet}.
\begin{table}
\centering
\begin{tabular} {|l|c|c|c|c|c|} \hline
layer & type & filter size & filters & stride   & activation \\  \hline
1 & conv & 3$\times$3 & 71 & 1  & tanh \\ 
2 & pool & 2$\times$2 & 1 & 2 &  max \\ 
3 & conv & 2$\times$2 & 88 & 1  & tanh \\ 
4 & pool & 2$\times$2 & 1 & 2 &  max \\ 
5 & fc & 1$\times$5 & 26 & 1 & tanh \\
6 & fc & 1$\times$1 & 5 &1 & softmax \\ \hline
\end{tabular}
\caption{Architecture of the supervised CNN}\
\label{table:convnet}
\end{table}
We trained the network on 45,000 examples using stochastic gradient descent. We then classified 15,000 test examples using the trained model. 
The classification performance was compared with k-nearest neighbor classifiers and support vector machines.

\subsection{Unsupervised Learning with Convolutional Autoencoders}
  \begin{table}
\centering
\begin{tabular} {|l|c|c|c|c|c|c|} \hline
layer & type & filter size & filters & stride &pad  & activation \\  \hline
1 & conv & 5$\times$5 & 16 & 1 & 2x2  & RELU \cite{alexnet} \\ 
2 & pool & 2$\times$2 & 1 & 2 & 0 &  max \\ 
3 & conv & 3$\times$3 & 16 & 1 & 1$\times$0  & RELU \\ 
4 & pool & 2$\times$2 & 1 & 2 & 0 & max \\ 
5 & fc & 2$\times$5 & 10 & 1 & 0 & RELU \\
6 & deconv & 2$\times$4 & 16 &2 & 0 & None\\ 
7 & deconv & 2$\times$5 & 16 &2 & 0 & None\\
8 & deconv & 2$\times$4 & 1 &2 & 0 & None \\ \hline
\end{tabular}
\caption{Architecture of the convolutional autoencoder}\
\label{table:convae}
\end{table}
For the convolutional autoencoder, we use the architecture specified in Table \ref{table:convae}, using sum of squared error as the loss function. The convolutional autoencoder was trained using gradient descent with a learning rate of 0.0005 and a momentum coefficient of 0.9. We trained the network on 31,700 training examples and tested it on 7900 test examples. 

\section{Results} 
\subsection{Supervised Learning with CNN}
\subsubsection{Results}
\begin{table}
\centering
\begin{tabular}{|c|ccccc|}
\hline
{Measure} & IBD & IBD & \multirow{2}{*}{Muon}   & \multirow{2}{*}{Flasher} & \multirow{2}{*}{Other}  \\
and Method & prompt & delay & & & \\\hline
$F_1$-score & & & & & \\
k-NN   & 0.775   & 0.954    & 0.996 & 0.784  & 0.806 \\
 SVM    & 0.846   & 0.962    & 0.996 & 0.895  & 0.885 \\
 CNN    & \textbf{0.891}   & \textbf{0.974}    & \textbf{0.997} & \textbf{0.951}  & \textbf{0.928} \\ \hline
 Accuracy & & & & & \\
 k-NN   & 0.950   & 0.990    & 0.998 & 0.891  & 0.896 \\
 SVM    & 0.966   & 0.992    & 0.998 & 0.947  & 0.938 \\
 CNN    & \textbf{0.977}   & \textbf{0.995}    & \textbf{0.999} & \textbf{0.974}  & \textbf{0.962} \\ \hline
\end{tabular}
\caption{Classifier performance for different events}
\label{table:classwise}
\end{table}

The classification classwise $F_1$-scores and classification accuracies  of k-nearest neighbor, support vector machine, and the CNN architecture on the test set are summarized in Table \ref{table:classwise}.
We also used t-SNE \cite{tsne} to visualize the features learned for the supervised convolutional neural network. Figure \ref{fig:supervised_tsne_3} shows the t-SNE visualization of the outputs from the last fully connected layer of the CNN. This visualization shows in two dimensions how the each example is clustered in the 26-dimensional feature space learned by the network. 

\begin{figure*}
  \centering
  \begin{subfigure}[b]{0.495\textwidth}
    \centering
    \includegraphics[width=0.9\linewidth,trim=2.9cm 2.8cm 2.7cm 6.5cm,clip]{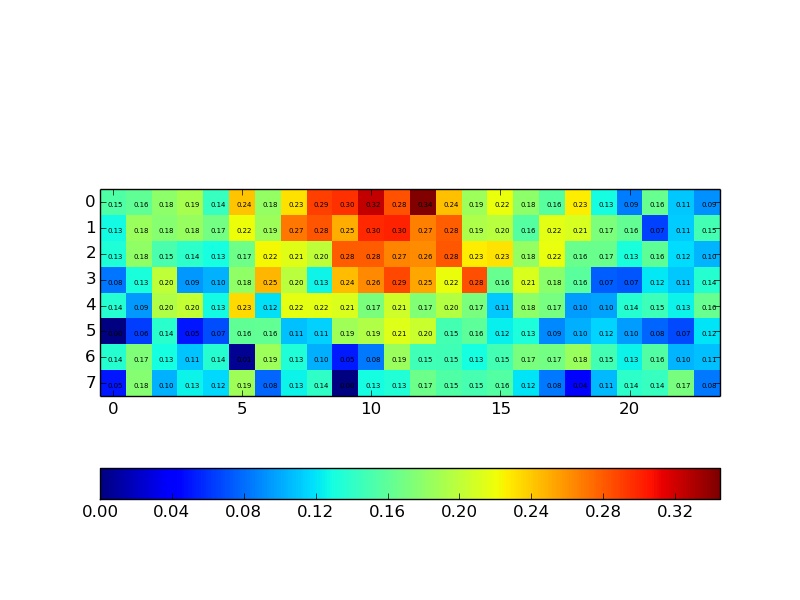}
    \caption{An IBD delay event in cluster A}
    \label{fig:delay-in-delay-prompt}
  \end{subfigure}
  \hfill
  \begin{subfigure}[b]{0.495\textwidth}
    \centering
    \includegraphics[width=0.9\linewidth,trim=2.9cm 2.8cm 2.7cm 6.5cm,clip]{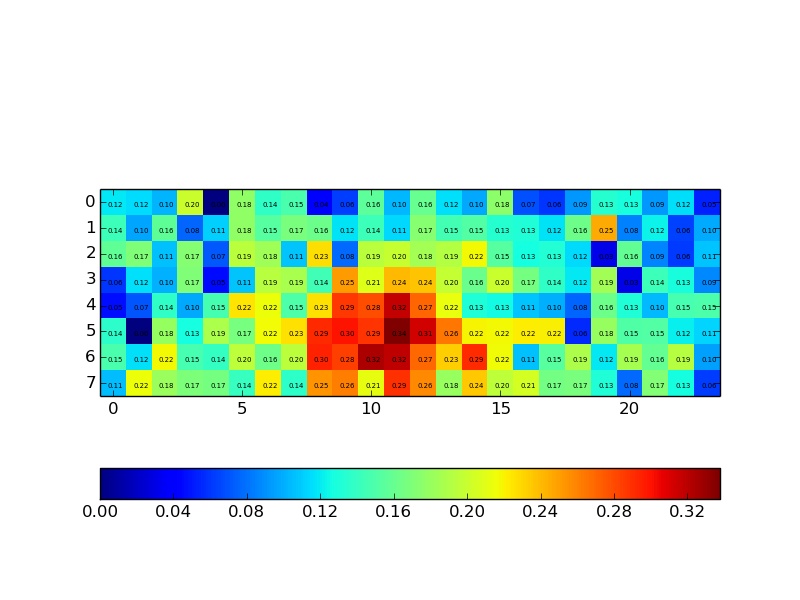}
    \caption{An IBD prompt event in cluster A}
    \label{fig:prompt-in-delay-prompt}
  \end{subfigure}
  \begin{subfigure}[b]{0.495\textwidth}
    \centering
    \includegraphics[width=0.9\linewidth,trim=2.9cm 2.8cm 2.7cm 6.5cm,clip]{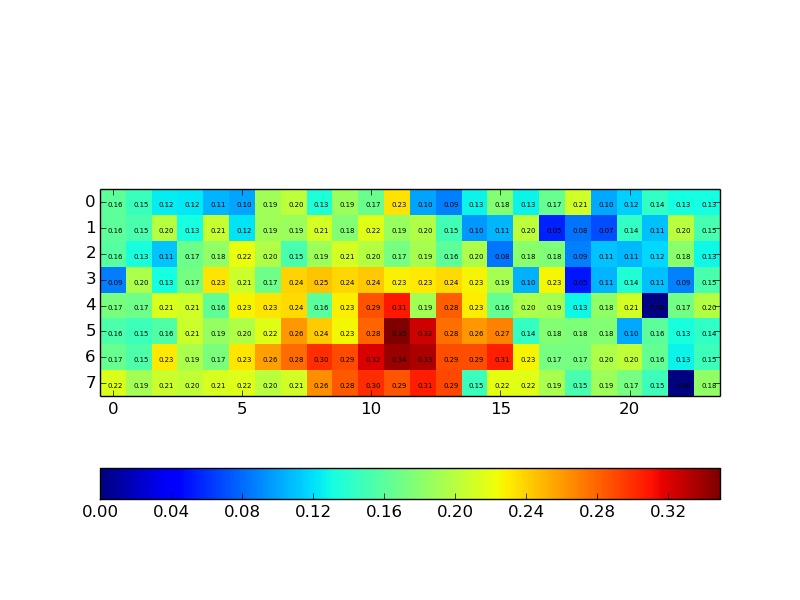}
    \caption{An IBD delay event in cluster C}
    \label{fig:delay-in-delay-delay}
  \end{subfigure}
  \begin{subfigure}[b]{0.495\textwidth}
    \centering
    \includegraphics[width=0.9\linewidth,trim=2.9cm 2.8cm 2.7cm 6.5cm,clip]{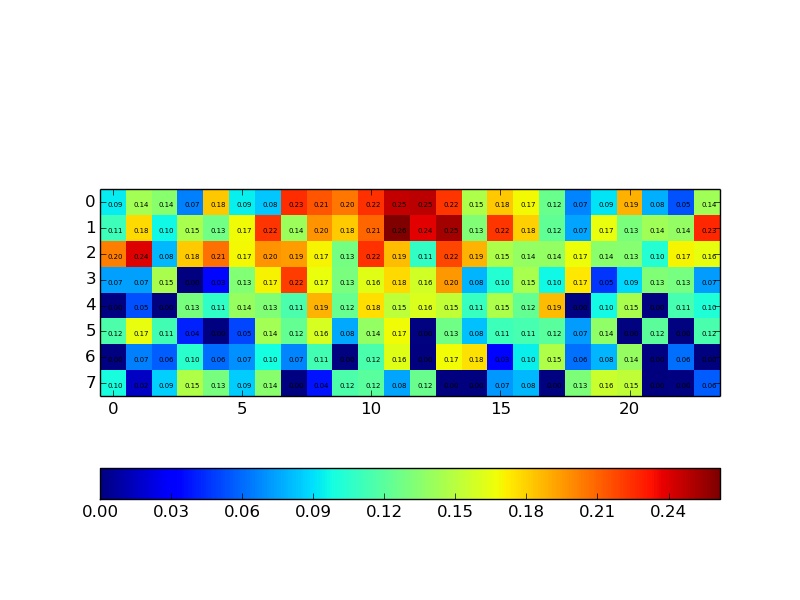}
    \caption{An IBD prompt event in the blue cluster below the letter B}
    \label{fig:prompt-in-prompt-prompt}
  \end{subfigure}
  \caption{Representative examples of various IBD events in the clusters labeled in Figure \ref{fig:supervised_tsne_3}.}\label{fig:ibd-events}
\end{figure*}
\begin{figure}
\centering
\includegraphics[width=0.95\linewidth,trim=2.5cm 1.6cm 2.2cm 1.8cm,clip]{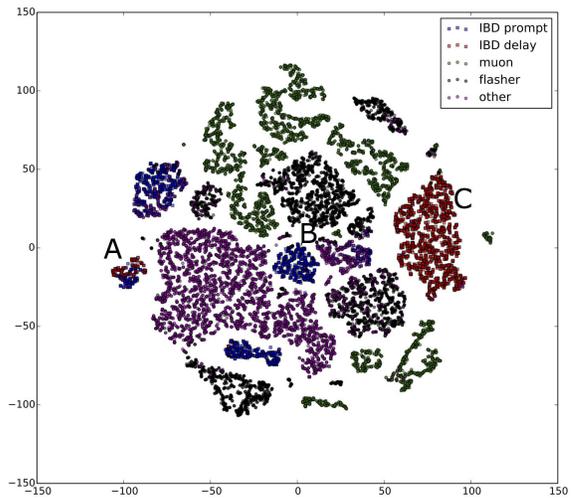}
\caption{t-SNE reduction of representation learned on the last fully connected layer of CNN. Representative examples from the clusters immediately below the labels A and B and to the left of C are shown in figure \ref{fig:delay-in-delay-prompt}--\ref{fig:prompt-in-prompt-prompt}}
\label{fig:supervised_tsne_3}
\end{figure}
We also show, in Figures \ref{fig:delay-in-delay-prompt} and \ref{fig:prompt-in-delay-prompt}, example PMT charges of different types of events that are in clusters in the t-SNE clustering (Figure \ref{fig:supervised_tsne_3}) that contain a mix of labels near each other, as well as examples contained in well separated clusters in Figures \ref{fig:delay-in-delay-delay} and \ref{fig:prompt-in-prompt-prompt}. These examples are visualizations of the 8x24 arrays after preprocessing. As described in the preprocessing section, the value of each element in the array is the raw charge deposit as measured by the PMT at the time of the trigger transformed by a natural log and then divided by a scale factor of 10 to ensure values between 0 and 1.

\subsubsection{Interpretation}
Our results suggest that there are patterns in the Daya Bay data that can be uncovered by machine learning techniques without knowledge of the underlying physics. Specifically, we were able to achieve high accuracy on classification of the Daya Bay events using only the spatial pattern of the charge deposits. In contrast, the physicists used the time of the events and prior physics knowledge to perform classification. In addition, our results suggest that deep neural networks were better than other techniques at classifying the images and thus finding patterns in the data. 
as shown in Table \ref{table:classwise}. Our CNN architecture had the highest $F_1$-score and accuracy for all event types. In particular, it showed significantly higher performance on classes ``IBD prompt'' and ``flasher''. 
Not only did the supervised CNN perform better in classifying the data then other shallower ML techniques, but it also discovered features in the data that helped cluster it into fairly distinct groups as shown in Figure \ref{fig:supervised_tsne_3}. 

\begin{figure}[t]
\centering
\includegraphics[width=\linewidth,trim=2.1cm 1cm 2.2cm 1.8cm,clip]{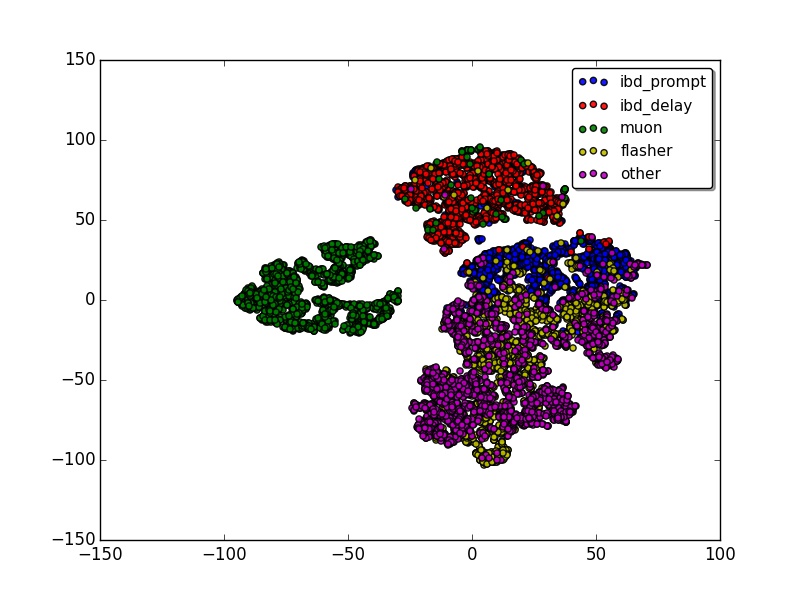}
\caption{t-SNE representation of features learned by the convolutional autoencoder}
\label{fig:convae-tsne2}
\end{figure}
\begin{figure*}
  \centering
  \begin{subfigure}[b]{0.495\textwidth}
    \centering
    \includegraphics[width=\linewidth,trim=3cm 1cm 2cm 2cm,clip]{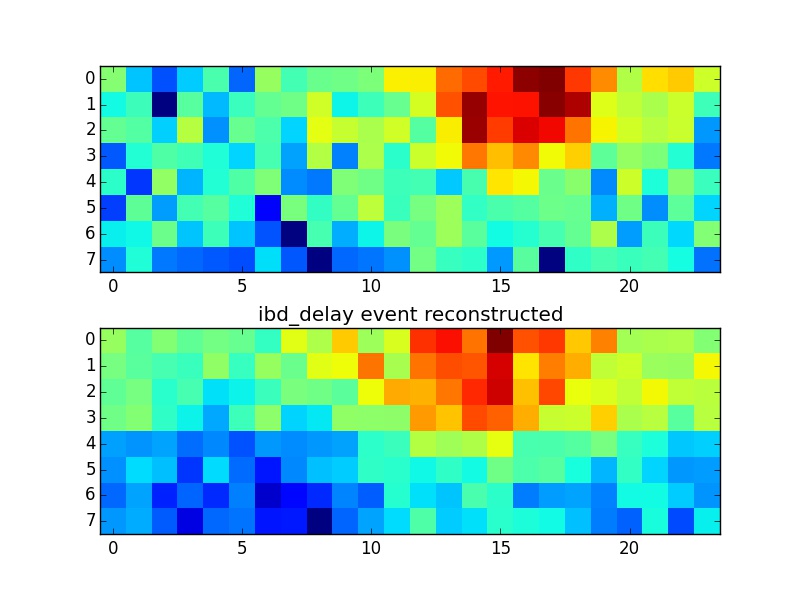}
    \caption{Example of an ``IBD delay'' event}
    \label{fig:delay}
  \end{subfigure}
  \begin{subfigure}[b]{0.495\textwidth}
    \centering
    \includegraphics[width=\linewidth,trim=3cm 1cm 2cm 2cm,clip]{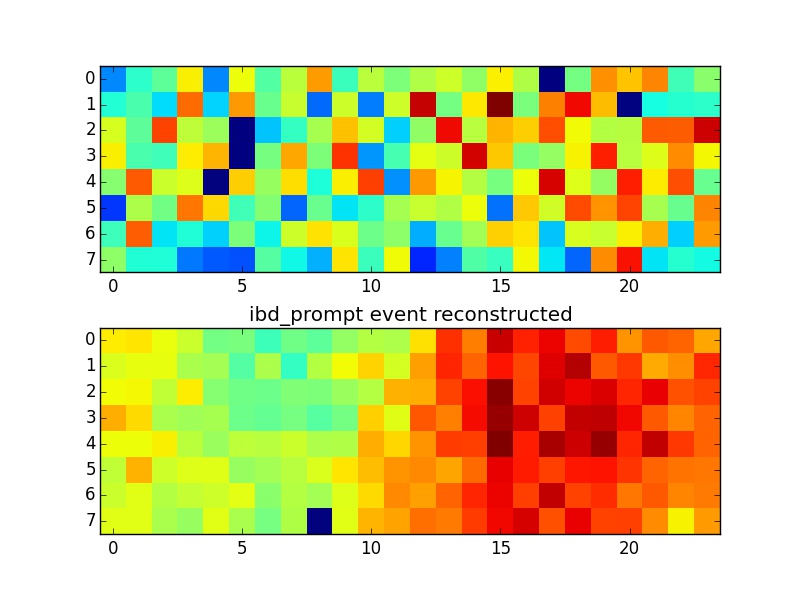}
    \caption{Example of an ``IBD prompt'' event}
    \label{fig:prompt}
  \end{subfigure}
  \caption{Raw event image (top row) and convolutional autoencoder reconstructed event image (bottom row).}
\end{figure*}

We can further investigate the raw images within the clusters formed by t-SNE. For example, in Figures \ref{fig:delay-in-delay-prompt} and  \ref{fig:prompt-in-delay-prompt} the CNN has identified a particularly distinctive charge pattern common to both images. Specifically, both images have the same range of values and have a very similar shape. Though the patterns happen at different parts of the image, they are roughly the same and it is not surprising that the CNN picked up on this translation invariant pattern. These are labeled as different types 
because prompt events have a large range of charge patterns, some of which very closely resemble delay events. The standard physics analysis is able to resolve these only by using the time coincidence of delay events happening within 200 microseconds after prompt events, while the neural network solely has charge pattern information. Future work involving these features may help solve this, but it is nevertheless encouraging that the network was able to hone in on the geometric pattern. Figures \ref{fig:delay-in-delay-delay} and \ref{fig:prompt-in-prompt-prompt}, on the other hand, show images from more distinct prompt and delay clusters, respectively, illustrating that prompt events deposit less energy in the detector on average as shown by the different range of values in the two images.
Such clustering suggests that, with help from ground truth labeling, deep learning techniques can discover informative features and thus find structure in raw physics inputs. Because such patterns in the data exist and can be learned, this suggests that unsupervised learning also has the potential to discover these patterns without needing ground truth labeling.

 \subsection{Unsupervised learning with Convolutional Autoencoder}

\subsubsection{Results}
For the convolutional autoencoder, we present the t-SNE visualization of the 10 features learned by the network in figure \ref{fig:convae-tsne2}. To show how informative the feature vector that the network learned is, we also show several event images and their reconstruction by the autoencoder in Figures \ref{fig:delay} and \ref{fig:prompt}. More informative features that are learned correspond to more accurate reconstructions because the 10 features effectively give the network the ``ingredients'' it needs to the reconstruct the input 8x24 structure.

\subsubsection{Interpretation}

The convolutional autoencoder is designed to reconstruct PMT images and so it learns different features than the supervised CNN which is attempting to classify based on the training labels. Therefore, the t-SNE clustering for this part of the study (in Figure \ref{fig:convae-tsne2}) is quite different from that in the supervised section. Nevertheless, we were able to obtain well defined clusters without using any physics knowledge. Specifically, there is a very clearly separated cluster that can be identified with the labelled muons, and also a fairly clear separation between ``IBD delay'' and other events. We even achieve some separation between ``IBD prompt'' and ``other'' backgrounds which, as mentioned above, is mainly achieved in the default physics analysis only by incorporating additional information of the time between prompt and delayed events. 


By looking at the reconstructed images, we can see the autoencoder was able to filter out the input noise and reconstruct the important shape of different event types. For example, in Figure \ref{fig:delay}, the shape of the charge pattern is reconstructed extremely accurately, which shows that the 10 learned features from the autoencoder are very informative for ``IBD delay'' events. In Figure \ref{fig:prompt}, salient and distinct aspects, like the high charge regions on the right side and the low regions on the left, of the more challenging ``IBD prompt'' events are also reconstructed well. 
As further work, it would be desirable to obtain better separation between ``flasher'' and ``other'' events. Therefore we intend to continue to tailor the convolutional autoencoder approach to this application by considering input transformations that take into account the experiment geometry, variable resolution images, and alternative construction of convolutional filters, as well as more  data and full parameter optimization of the number of filters and the size of the feature vector. 


%

\section{Conclusions} 
In this work we have applied for the first time unsupervised deep neural nets within particle physics and have shown that the network can successfully identify patterns of physics interest. As future work we are collaborating with physicists on the experiment to investigate in detail the various clusters formed by the representation to determine what interesting physics is captured in them beyond the initial labelling. We also plan to incorporate such visualizations into the monitoring pipeline of the experiment.


Such unsupervised techniques could be utilized in a generic manner for a wide variety of particle physics experiments and run directly on the raw data pipeline to aid in trigger (filter) decisions or in evaluating data quality, or to discover new instrument anomalies (such as flasher events). The use of unsupervised learning to identify such features is of considerable interest within the field as it can potentially save considerable time required to hand-engineer features to identify such anomalies.

We have also demonstrated the superiority of convolutional neural networks compared to other supervised machine learning approaches for running directly on raw particle physics instrument data. This offers the potential for use as fast selection filters, particularly for other particle physics experiments that have many more channels and approach exabytes of raw data such as those at the current Large Hadron Collider (LHC) and planned HL-LHC at CERN \cite{AtlasHLLHCLoI}. Our analysis in this paper used the labels determined from an existing physics analysis and therefore the selection accuracy is upper bounded by that of the physics analysis. Many other particle physics experiments, however, have reliable simulated data which could be used with the approaches in this paper to better the selection accuracy achieved with those experiments' current analyses. 


In conclusion, we have demonstrated how deep learning can be applied to reveal physics directly from raw instrument data even with unsupervised approaches, and therefore that these techniques offer considerable potential to aid the fundamental discoveries of future particle physics experiments.

\section*{Acknowledgments}

The authors gratefully acknowledge the Daya Bay Collaboration for access to their experimental data and many useful discussions, and specifically Yasuhiro Nakajima for the dataset labels, and physics background details. 

This research was conducted using neon, an open source library for deep learning from Nervana Systems. This research used resources of the National Energy Research
Scientific Computing Center, a DOE Office of Science User Facility 
supported by the Office of Science of the U.S. Department of Energy 
under Contract No. DE-AC02-05CH11231. 

This work was supported by the Director, Office of Science,
Office of Advanced Scientific Computing Research,
Applied Mathematics program of the U.S. Department of
Energy under Contract No. DE-AC02-05CH11231.

S. Ko was supported by Basic Science Research Program through the National Research Foundation of Korea (NRF) grants funded by the Korea government (MSIP) \\                                 (Nos. 2013R1A1A1057949 and 2014R1A4A1007895).

\bibliographystyle{IEEEabrv}
\bibliography{refs}

\end{document}